\documentclass{article}

\usepackage[preprint]{corl_2026} 
\usepackage{graphicx}
\usepackage[table,xcdraw,dvipsnames]{xcolor}
\usepackage[T1]{fontenc}
\usepackage{epsfig}
\usepackage{amsmath}
\usepackage[capitalise]{cleveref}
\usepackage{amssymb}
\usepackage{color, soul, colortbl}
\usepackage{mathtools}
\usepackage{booktabs}
\usepackage{times}
\usepackage{microtype}
\usepackage[labelfont=bf, font=footnotesize]{caption}
\usepackage[labelfont=bf, font=footnotesize]{subcaption}
\usepackage[utf8]{inputenc}
\usepackage{float}
\usepackage{inconsolata}
\usepackage{multirow}
\usepackage{placeins}
\usepackage{stfloats}
\usepackage{enumitem}
\usepackage{tabularx}
\usepackage{xstring}
\usepackage{multirow}
\usepackage{xspace}
\usepackage{url}
\usepackage{hyperref}
\usepackage{duckuments}
\usepackage{svg}
\usepackage[hang,flushmargin]{footmisc}
\usepackage{url}
\usepackage{tabularray}
\usepackage{wrapfig}
\usepackage{epigraph}
\usepackage{amssymb}

\newcommand{\R}[1]{{%
    \textbf{%
        \ifstrequal{#1}{1}{\textcolor{red}{R#1}}{%
        \ifstrequal{#1}{2}{\textcolor{blue}{R#1}}{%
        \ifstrequal{#1}{3}{\textcolor{magenta}{R#1}}{%
        \ifstrequal{#1}{4}{\textcolor{teal}{R#1}}{%
                           \textcolor{cyan}{R#1}%
        }}}}%
    }%
}}

\definecolor{tabfirst}{rgb}{0.7, 1.0, 0.7} 
\definecolor{tabsecond}{rgb}{1, 1, 0.7} 
\definecolor{tabthird}{rgb}{1, 0.85, 0.7} 

\newcommand{\raisemath}[1]{\mathpalette{\raisemath{#1}}}

\newcommand{\Ac}{\mathcal{A}}

\newcommand{\Lc}{\mathcal{L}}

\newcommand{\Pc}{\mathcal{P}}

\newcommand{\Rc}{\mathcal{R}}
\newcommand{\Sc}{\mathcal{S}}

\newcommand{\Xc}{\mathcal{X}}

\newcommand{\Eb}{\mathbb{E}}

\newcommand{\Nb}{\mathbb{N}}

\newcommand{\Rb}{\mathbb{R}}

\newcommand{\Ev}{\mathbf{E}}

\newcommand{\Mv}{\mathbf{M}}

\newcommand{\Rv}{\mathbf{R}}

\newcommand{\Tv}{\mathbf{T}}

\newcommand{\norm}[1]{\left\lVert#1\right\rVert}

\definecolor{RebuttalColor}{RGB}{0,0,0}

\DeclareMathOperator*{\argmax}{arg\,max}

\definecolor{pmcolor}{RGB}{70,70,70}
\newcommand{\pmcolor}[1]{\textcolor{pmcolor}{#1}}
\newcommand{\pms}[2]{#1{\tiny{{\pmcolor{{$\pm$#2}}}}}}

\newcommand{\ourmethod}{\texttt{PTLD}\xspace}

\title{PTLD: Sim-to-Real Privileged Tactile Latent Distillation for Dexterous Manipulation}

\author{
  Rosy Chen\\
  Carnegie Mellon University \\
  \And
  Mustafa Mukadam \\
  University of Washington \\
  \And
  Michael Kaess\\
  Carnegie Mellon University \\
  \And
  Tingfan Wu \\
  FAIR at Meta \\
  \And
  Francois R Hogan \\
  FAIR at Meta \\
  \And
  Jitendra Malik \\
  UC Berkeley
  \And
  Akash Sharma \\
  Carnegie Mellon University \\
  \texttt{\{dingdinc, akashsha\}@andrew.cmu.edu}
}

\begin{document}
\maketitle
\begin{figure}[htbp]
    \centering
    \includegraphics[width=0.75\linewidth]{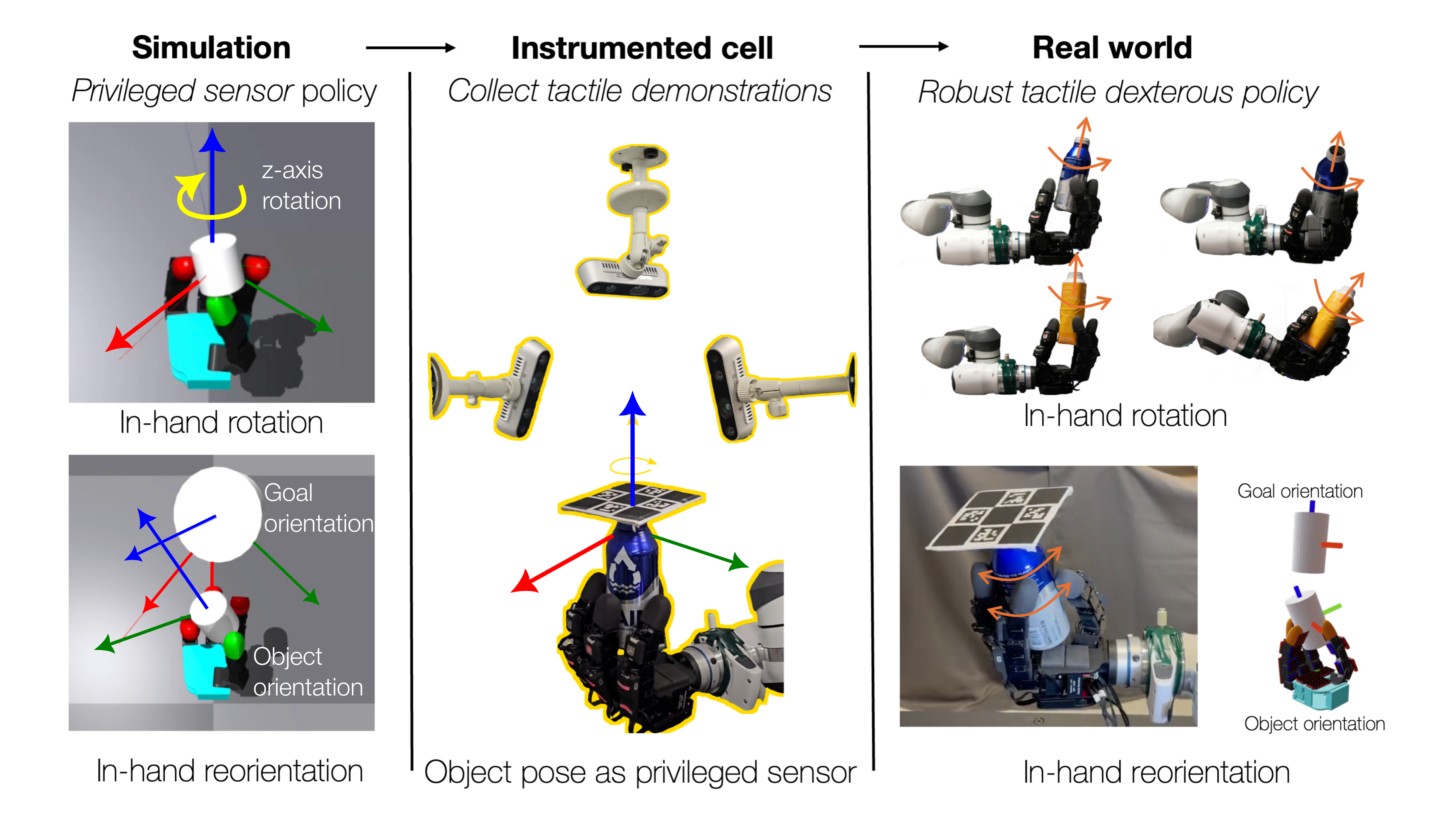}

    \caption{\ourmethod trains tactile dexterous policies without simulating tactile sensors. First, \emph{privileged sensor} policies are trained in simulation via reinforcement learning (RL), then deployed in a real-world instrumented cell to collect tactile demonstration. Finally, a tactile encoder is trained on the demonstrations to mimic the \emph{privileged sensor} encoder. At deployment, the tactile policy relies solely on tactile signals and proprioception, without external pose tracking, making it robust to object property changes, wrist orientation changes, and generalizing to irregular geometries such as cube.} 
    \vspace{-0.5cm}
    \label{fig:teaser}
\end{figure}

\begin{abstract}\label{sec:abstract}
Tactile dexterous manipulation is essential to human manipulation, yet learning effective tactile enabled robot policies remains a challenge.
While recent work relies on imitation learning, obtaining high quality demonstrations for multi-fingered hands via robot tele-operation or kinesthetic teaching is prohibitive.
Alternatively, with reinforcement learning (RL) we can learn skills in simulation, but the lack of fast and realistic tactile simulation remains a core issue. To bridge this gap, we introduce sim-to-real Privileged Tactile Latent Distillation (\ourmethod). Our method anchors sim-to-real transfer on privileged sensors and learns a tactile encoder that maps real-world touch into the latent state of the privileged simulation policy, bypassing tactile simulation entirely.  
We demonstrate that \ourmethod improves over proprioceptive manipulation policies trained in simulation significantly by incorporating tactile sensing. On the benchmark in-hand rotation task, \ourmethod achieves a 182\% improvement. \ourmethod also enables the challenging task of tactile in-hand reorientation where we see a 57\% improvement in number of goals reached. Website: \url{https://akashsharma02.github.io/ptld-website/}
\end{abstract}\vspace{-1em}

\section{Introduction}\label{sec:introduction}

Human-like dexterous manipulation remains a fundamental challenge in robotics. It is critical for deploying physical intelligence in unstructured environments like healthcare and homes. While recent work in learning from demonstrations (LfD)~\cite{chi2023diffusionpolicy, zhao2023learning, chi2024universal} provides a scalable recipe to learn policies from large demonstration datasets collected via teleoperation~\cite{zhao2023learning}, hand-held grippers~\cite{chi2024universal, xu2025dexumi}, or kinesthetic teaching, dexterous hand hardware poses a unique challenge. Reliable teleoperation for intricate tasks like using a screwdriver, a wrench, or turning a doorknob is 
difficult. Kinesthetic teaching is equally challenging when more than two fingers are required for tasks such as in-hand object reorientation~\cite{chen2025dexforce}. Hand-held grippers are promising~\cite{chi2024universal, xu2025dexumi} but, they require exoskeleton structures that balance flexibility and stability; consequently, recent successes have been limited to simple tasks.

Sim-to-real RL offers an alternative for learning dexterous tasks and has seen significant success in robot locomotion~\cite{tan2018sim, peng2018deepmimic}. However, most existing approaches focus on blind, proprioception-only policies for both locomotion~\cite{he2025hover, li2025bfm, zakka2025mujoco} and manipulation~\cite{qi2022hand, pitz2023dextrous}. While a few works have succeeded with perceptive policies~\cite{handa2023dextreme, singh2024dextrah, akkaya2019solving}, training visual policies in simulation is slow due to the overhead of image rendering. Furthermore, these policies are often challenged by a significant sim-to-real gap.

Our focus in this work is on \emph{tactile dexterous manipulation}, encompassing \emph{dynamic} tasks such as in-hand rotation, and reorientation~\cite{qi2022hand, andrychowicz2020learning, higuera2025tactile, yang2024anyrotate, yin2025dexteritygen}. The standard approach for these tasks is RL.   
However, two core challenges make training effective tactile policies in simulation difficult.
First, simulating tactile sensors accurately is difficult. Thus, existing works~\cite{yang2024anyrotate, yin2023rotating} resort to simplified  contact models such as single-point~\cite{qi2023general} or binary contact~\cite{yin2024learninginhandtranslation}, which fail to capture the rich force and geometry information that real tactile sensors provide. Furthermore, existing simulation packages~\cite{wang2022tacto, akinola2025tacsl, su2024sim2real} are not standardized for all tactile sensors and primarily rely on rigid-body simulation, compounding the accuracy problem. Second, the sim-to-real gap for tactile sensing remains large - real-world contact dynamics, sensor hysteresis, and surface properties are difficult to reproduce in simulation, preventing straightforward deployment of simulation-trained tactile policies.

We present \ourmethod, a new approach to learning tactile manipulation policies without paying the high cost of its simulation. 
Our method takes inspiration from \emph{privileged latent distillation}~\cite{chen2020learning, choudhury2018data, kumar2021rma}, where an `oracle' policy is first trained with access to privileged state and then imitated by a deployable policy using only partial observations. We extend privileged latent distillation in two distinct ways. First, we cast sim-to-real transfer as finding a shared latent space between simulation and the real world — one in which tactile observations in real recovers privileged policy state used in simulation. Second, we treat properties such as object pose and shape as \emph{privileged sensors} and deploy these policies in the real world by instrumenting a robot cell (\cref{fig:teaser}). 
On deployment, we record paired (tactile observation,  privileged sensor latent) tuples and train an encoder. The result is a deployable tactile policy with internal representation aligned, by construction, with the \emph{privileged sensor} policy. 

Because \ourmethod relaxes distillation to leverage \emph{privileged sensors}, it requires an additional round of real-world distillation. This can make two-stage simulation training laborious. Therefore, in this work, we also present \emph{online latent distillation} for Asymmetric Actor-Critic (AAC)~\cite{pinto2017asymmetric} and demonstrate that it can outperform two-stage distillation. 

In summary, our contributions are threefold: 
\begin{itemize}[itemsep=-2pt,topsep=-2pt,leftmargin=7mm]
    \item We present a novel approach to learn tactile dexterous manipulation policies without paying the cost of its simulation. We use privileged sensors as the interface between simulation and reality to perform \emph{privileged tactile latent distillation} (\ourmethod).
    \item We present \emph{online latent distillation} for AAC, and simplify two-stage distillation in simulation into a single training step.  
    \item We show that \ourmethod tactile policies for continuous in-hand rotation and in-hand reorientation - a task that cannot be trained in simulation using proprioceptive history alone - outperform proprioception and adaptation based tactile policies in both robustness and performance. 
\end{itemize}
\section{Related work}\label{sec:related_work}
\paragraph{Dexterous In-hand Manipulation}
Dexterous in-hand manipulation has been an active area of research for decades~\cite{okamura2000overview, akkaya2019solving, andrychowicz2020learning, rus1999hand, nagabandi2020deep, han1998dextrous}.
While classical approaches need a physical model of the object and robot geometry to plan robot finger motions~\cite{fearing1986implementing, morgan2022complex}, recent approaches have had success with using RL directly to learn policies in a model free manner~\cite{akkaya2019solving, andrychowicz2020learning, chen2023visual, qi2022hand, yin2023rotating}. However, RL approaches face the \emph{sim-to-real} gap. i.e. it is challenging to reproduce real world sensor observation and physics in simulation. Even for modalities such as vision where simulation is feasible, the model is often physically inaccurate failing to describe real world sensor properties, therefore extensive visual domain randomization~\cite{tobin2017domain, handa2023dextreme} is crucial. Our method, on the other hand trains in simulation with observations such as object poses and object shape that typically do not suffer a large sim-to-real gap, but requires one to forgo the zero-shot sim-to-real deployment assumption. 
\paragraph{Tactile sensing and Representation learning} 
While the tactile modality has long been promised to be imperative for contact-rich dexterous manipulation, its application has historically been limited. Despite a decade of innovation in hardware—including vision-based (e.g., GelSight~\cite{yuan2017gelsight}, DIGIT~\cite{lambeta2020digit}), magnetic (e.g., ReSkin~\cite{bhirangi2021reskin}, Xela~\cite{tomo2018xela}), and resistive and capacitive~\cite{huang3d} sensors—tactile-driven manipulation remains largely confined to simple, quasi-static tasks like peg-insertion~\cite{sharma2025selfsupervised, dong2021icra}, cable manipulation~\cite{she2021cable} or planar pushing~\cite{suresh2021tactile}. These tasks are often selected specifically for their suitability for behavior cloning. 
On the other hand, recent self-supervised representations such as Sparsh~\cite{higuera2024sparsh, sharma2025selfsupervised, higuera2025tactile, zhao2024transferable} have begun to address sensor standardization demonstrating their use in several manipulation tasks, yet the tasks chosen are largely still quasi-static. Furthermore, while tactile adaptation~\cite{higuera2025tactile, wang2024penspin} algorithms have attempted to refine RL policies using real-world data, these methods are fundamentally constrained by the performance ceilings of their proprioceptive teacher policies and often rely on simple rejection sampling. In contrast, \ourmethod learns quantitatively more robust and dynamic policy behaviors from \emph{privileged sensor} teacher policies.
\section{Background}\label{sec:background}\vspace{-0.5em}
\paragraph{Notation}
We model the tasks discussed in this paper as finite horizon ($N \in \Nb$) Partially Observable Markov Decision Processes (POMDPs) $\Mv \triangleq (\Sc, \Ac, \Xc, \Pc, \Rc)$ where $(\Sc, \Ac, \Xc) \in \{\Sc_t, \Ac_t, \Xc_t\}_{t=1}^N$ denote the state, action and observation spaces over the finite horizon $N$ respectively. $\Pc = \{\Pc_t: \Sc_{t-1} \times \Ac_{t-1} \rightarrow \Sc_t\}$ denotes the transition dynamics, and $\Rc = \{\Rc_t: \Sc_t \times \Ac_t \rightarrow [0, 1]\}$ denotes the reward function. Our goal is to learn policies $\pi: \Xc_{t-k+1:t} \times \Ac_{t-k:t-1} \rightarrow \Ac_{t}$ via Reinforcement Learning (RL) to maximize the expected return $G(\tau) = \sum_{t=0}^{N} \gamma^t R_t$  over the horizon as $\pi^* = \argmax_{\tau \sim \Pc^\pi} \Eb\left[ G(\tau) \right].$

As in~\cref{fig:rma_vs_aac}, our network parameterization is composed of an encoder $\Ev$ which encodes the observations $\Xc$ into a latent space $\Lc$, and a policy $\pi$ to produce actions $\Ac$. Typically, we employ two encoders during training. First, we have $\hat{\Ev}$ the privileged encoder which has access to privileged observations in simulation, and $\Ev$ the adaptation encoder which only has access to deployable observations. Then we have the policy as follows: $\Ac_t \sim \pi(\Ev(\Xc_{t-k+1:t}), \Ac_{t-k:t-1})$\vspace{-1em}
\subsection{Privileged latent distillation}
\begin{figure*}[!t]
    \centering
    \includegraphics[width=\linewidth]{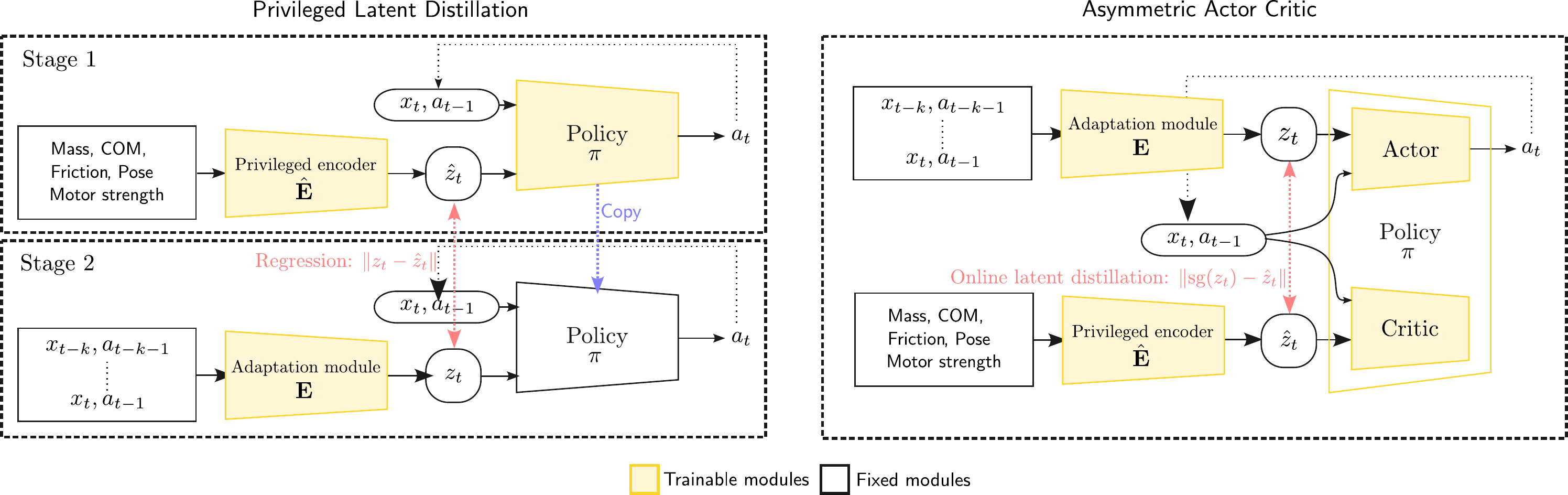}
    \caption{\textbf{(left)} \emph{Privileged latent distillation} is a two stage approach to training policies in simulation. An \emph{oracle} policy with privileged information is trained in stage 1, then it is distilled into a deployable policy in stage 2 (in simulation). \textbf{(right)} Asymmetric Actor Critic is a single stage approach where the actor and critic networks are trained simultaneously. The critic is provided with privileged information and learns the value function, while the actor is only given deployable partial sensor information}
    \label{fig:rma_vs_aac}\vspace{-1em}
\end{figure*}

\vspace{-0.5em}
Privileged latent distillation is a two stage approach for training deployable policies (\cref{fig:rma_vs_aac}). It requires simulation support for both privileged states $\Xc^\text{priv}$ (e.g., object and contact states) and real-world sensor observations $\Xc^\text{sensor}$ such as proprioception and vision. First, an oracle policy that is allowed to `cheat' is trained via RL (e.g., PPO) with full access to $\Xc^\text{priv}$. Subsequently, a student policy $\pi$ is trained to imitate the oracle using a history of sensor observations. Distillation typically minimizes either action or latent imitation loss: 
\begin{align*}
    \Lc_\text{action} = \norm{\hat{\pi}(\hat{\Ev}(\Xc^\text{priv}), \Ac) - \pi(\Ev(\Xc^\text{sensor}), \Ac)}, \qquad
    \Lc_\text{latent} = \norm{\hat{\Ev}(\Xc^\text{priv}) - \Ev(\Xc^\text{sensor})}.
\end{align*}
Experience for distillation is collected by the deployable (student) policy, while the supervision signal comes from the oracle encoder. This is importantly distinct from traditional offline imitation learning as student distillation implements an on-policy variant of DAgger~\cite{ross2011reduction}. Specifically, since the student policy is supervised by the teacher on experience collected by the student, the student observes a wider observation space during training, resulting in a robust policy.

\subsection{Asymmetric Actor Critic}\vspace{-0.5em}
Asymmetric Actor Critic (AAC)~\cite{pinto2017asymmetric} aims to learn robust deployable policies by taking advantage of full-state observability in simulation. Specifically, it employs an actor-critic framework, where the critic is provided with the privileged state $\Xc^\text{priv}$, while the actor is provided with $\Xc^\text{sensor}$ sensor observations (see \cref{fig:rma_vs_aac}). In our approach, we employ learning policies with AAC, as opposed to RMA~\cite{kumar2021rma} as it simplifies policy learning in simulation into a single training step. 
\section{\ourmethod: Privileged Tactile Latent Distillation}\label{sec:method} 
\subsection{Online latent distillation with Asymmetric Actor Critic}\label{subsec:method-online-distillation}
We employ an AAC framework for training manipulation policies in our work as opposed to privileged latent distillation, as it simplifies training into a single stage in simulation. 
Crucially, parameterizing the actor as an observation encoder $\Ev$ and policy $\pi$ is beneficial, as a distinct encoder allows learning general state representations. Therefore inspired by self-distillation in representation learning~\cite{grill2020bootstrap, baevski2022data2vec, sharma2025selfsupervised, higuera2025tactile} approaches, we employ a self-distillation representation loss between the latent representations learnt by the critic (privileged) encoder and the actor (student) encoder: 
\begin{align}
    \Lc_\text{latent} \triangleq \norm{ \Ev(\Xc^\text{sensor}) - \text{sg}(\hat{\Ev}(\Xc^\text{priv})) }
\end{align}
where $\text{sg}$ denotes stop gradient. This loss encourages the actor encoder to recover privileged object information available to the critic while operating on partial observations. In experiments (\cref{subsec:sim_results}), online latent distillation loss improves training reward achieved by the policy. Furthermore, in simulation evaluation, incorporating this simple distillation loss results in similar performance to privileged latent distillation.  

As shown in~\cref{fig:rma_vs_aac}, these networks are trained simultaneously using the clip variant of proximal policy optimization (PPO)~\cite{schulman2017proximal} augmented with the online latent distillation loss: 
\begin{align}
    \Lc_\text{PPO} &\triangleq \Lc_\pi^\text{CLIP}(\Ev, \pi) + c_V \Lc_V(\hat{\Ev}, V) + \Lc_\text{entropy} (\Ev, \pi) \\
    \Lc &\triangleq \Lc_\text{PPO} + c_\text{latent}\Lc_\text{latent}
\end{align}
where $c_\text{latent}$ is a weighting factor. We optimize the total loss $\Lc$ via backpropagation.

\subsection{Privileged sensors for Sim-to-Real Tactile Distillation}
\begin{figure*}[t]
    \centering
    \begin{minipage}[t]{0.48\textwidth}
        \centering
        \includegraphics[width=\linewidth]{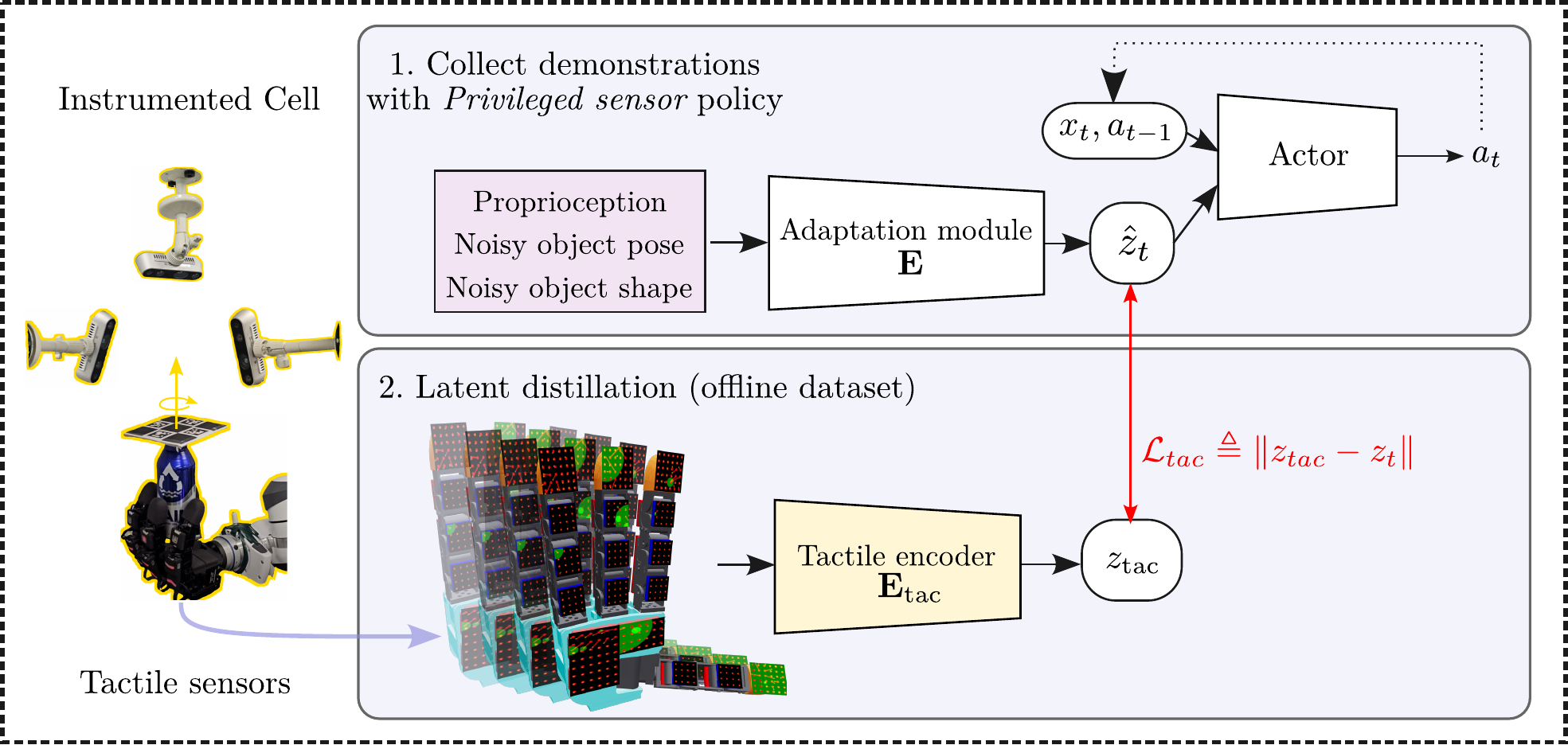}
        \caption{\ourmethod: Once we have a \emph{privileged sensor} policy trained using AAC, we collect demonstrations with deployment sensor observations in the real world via policy rollouts. Then, we train a deployment encoder (tactile encoder) to recover latents from the \emph{privileged sensor} policy}\label{fig:privileged_tactile_latent_distillation}\vspace{-1em}
    \end{minipage}
    \hfill
    \begin{minipage}[t]{0.48\textwidth}
        \centering
        \includegraphics[width=\linewidth]{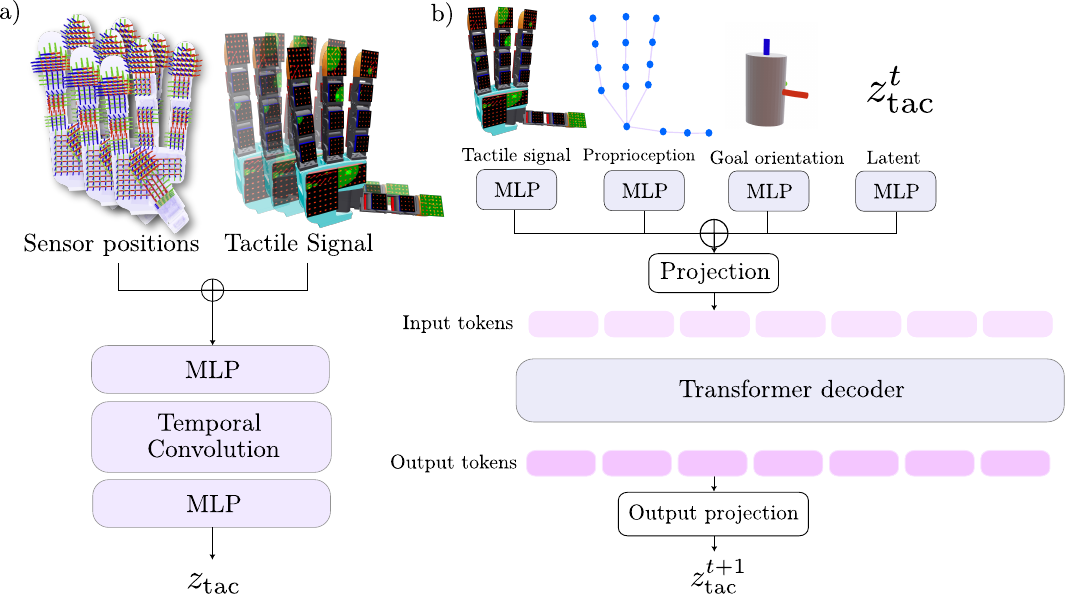}
        \caption{Tactile encoders: (a) For in-hand rotation, a 1D temporal convolutional network maps tactile and sensor position history to latents. (b) For in-hand reorientation, a causal transformer predicts future latents by embedding tactile signals, proprioception, goal orientations, and latent history.}
        \label{fig:tactile-encoder-architectures}\vspace{-2em}
    \end{minipage}
\end{figure*}

The essence of our method relies on the observation that by allowing the actor access to \emph{privileged sensors} that provide higher observability into the state, such as object pose and object shape, significantly improves policy performance in simulation compared to policies that take as input only partial observations $\Xc^\text{sensor}$ such as \emph{proprioception}.
Despite this gap, most prior work still uses the suboptimal policy at deployment time to achieve zero-shot sim-to-real transfer, sacrificing policy performance for simpler but deployable sensor modalities. In contrast, the \emph{privileged sensor} policy cannot be directly deployed in the real world as these observations are inaccessible at deployment time. To overcome this limitation, 
we instrument a real-world cell with multiple cameras and object markers to provide (noisy) object poses $\Tv_t^W \in \mathbf{SE}(3)$ as the real world \emph{privileged sensor}, and additionally also \emph{sensorize} the multi-fingered robot hand with tactile sensors. Then we deploy the \emph{privileged sensor} policy in the real-world cell and collect an offline dataset of \emph{on policy demonstrations}, recording both the latent representations produced by the policy and the associated tactile sensor observations. Using this data, we train an observation encoder that takes as input both tactile sensor data and proprioception data to match the latents from the privileged sensor policy. Formally, as illustrated in \cref{fig:privileged_tactile_latent_distillation} we distill the \emph{privileged sensor} policy into a tactile policy using MSE loss for supervision. At test-time deployment, the tactile encoder replace the \emph{privileged sensor} encoder and no privileged sensor information is used. 

Since the tactile encoder does not have access to a simulator for interactive training, naïve offline distillation can suffer from distribution shift. To address this, we employ DAgger~\cite{ross2011reduction}, where we iteratively train the tactile encoder with an aggregated dataset with experience collected by the policy using intermediate trained tactile encoders. 

\vspace{-1em}
\section{The privileged tactile manipulation system} \label{sec:system}\vspace{-1em}

For all experiments, we use the Allegro hand with Xela tactile skins and a calibrated multi-camera cell to provide privileged object pose. We provide additional hardware system details in the appendix.\vspace{-1em}

\subsection{Manipulation task I: In-hand rotation}
We choose the task of $z$-axis in-hand rotation~\cite{qi2022hand, qi2023general, yang2024anyrotate} for our experiments. In this task the robot hand is required to rotate an object along the $z$-axis, while ensuring the object remains held by the robot fingertips. We demonstrate that task performance on this task can be vastly improved by incorporating rich tactile information as part of the policy observation. For state, observation and reward details for this task, we refer the reader to~\cite{qi2022hand}. \vspace{-1em}
\paragraph{Tactile encoder:} For the tactile encoder (\cref{fig:tactile-encoder-architectures} (a)), we concatenate the tactile observations $\Xc^\text{tactile} \in \Rb^{368 \times 3}$, and the tactile sensor positions computed from the Allegro joint states using forward kinematics $\Xc^\text{sensor\_pos} \in \Rb^{368 \times 3}$ as the input $\Xc = \text{cat}(\Xc^\text{tactile}, \Xc^\text{sensor\_pos}) \in \Rb^{368 \times 6}$. The tactile observations are produced at 100Hz, and we use a history of 0.5s of tactile data as input to the encoder. Specifically, our tactile encoder uses a combination of MLP and 1D temporal convolution as follows: $(\Xc \rightarrow \text{MLP} \rightarrow \text{Temporal conv} \rightarrow \text{MLP} \rightarrow z)$, where $z$ is the predicted latent. 

\subsection{Manipulation task II: In-hand reorientation}\label{subsec:reorientation}
We further design an in-hand re-orientation task to show \ourmethod can not only improve real-world proprioceptive policies but also learn more complex information-heavy tasks, which are infeasible to learn in simulation without \emph{privileged sensors} (object pose). The goal is to reach an object pose $\Rv_t^\text{object}$ within $\delta \leq 0.25~\text{rad}$ of a target $\Rv_t^\text{goal}$. Constrained by hardware visibility, we sample goals within the upper hemisphere ($\theta = 40^\circ$ from the $z$-axis). We employ PPO with AAC for training. For reward and reset details, we refer the reader to the Appendix. \vspace{-1em}
\paragraph{State:} The adaptation module $\Ev$ receives Allegro joint states $q_t \in \Rb^{16}$, previous targets $\tilde{q}_t \in \Rb^{16}$, and noisy object pose $(p_t, \Rv_t) \in (\Rb^3, \mathbf{SO}(3))$ and goal orientations $\Rv_t^\text{goal} \in \mathbf{SO}(3))$. Orientations use 6-D representations \cite{zhou2019continuity}. We concatenate the relative pose difference to the goal to facilitate learning; notably, providing the current object orientation is essential for policy success. To capture temporal context, we provide a 30-step observation history ($\sim1.5$s at 20Hz) via frame stacking. The privileged encoder $\hat{\Ev}$ further receives object velocities and full fingertip states (poses and velocities).\vspace{-1em}

\paragraph{Tactile encoder} 
To handle the complexity of re-orientation and gait adaptation, we employ a \emph{recursive state estimator} as the tactile encoder. A causal Transformer (\cref{fig:tactile-encoder-architectures}b) auto-regressively predicts the next latent $z_{t+1}$ using history. We embed tactile observations $\Xc_t^\text{tactile} \in \Rb^{368 \times 3}$, proprioception $\Xc_t^\text{proprio} \in \Rb^{32}$, goal orientations $\Rv_t^\text{goal} \in \mathbf{SO}(3)$, and the previous latent $z_{t} \in \Rb^{8}$ via individual 2-layer MLPs. These embeddings are concatenated, projected with positional encodings, and processed by the Transformer decoder to infer the latent during inference. \vspace{-1em}
\section{Experiments}\label{sec:experiments}

\begin{figure*}[t!]
    \centering
    \begin{subfigure}[b]{0.48\textwidth}
        \centering
        \includegraphics[width=0.9\linewidth]{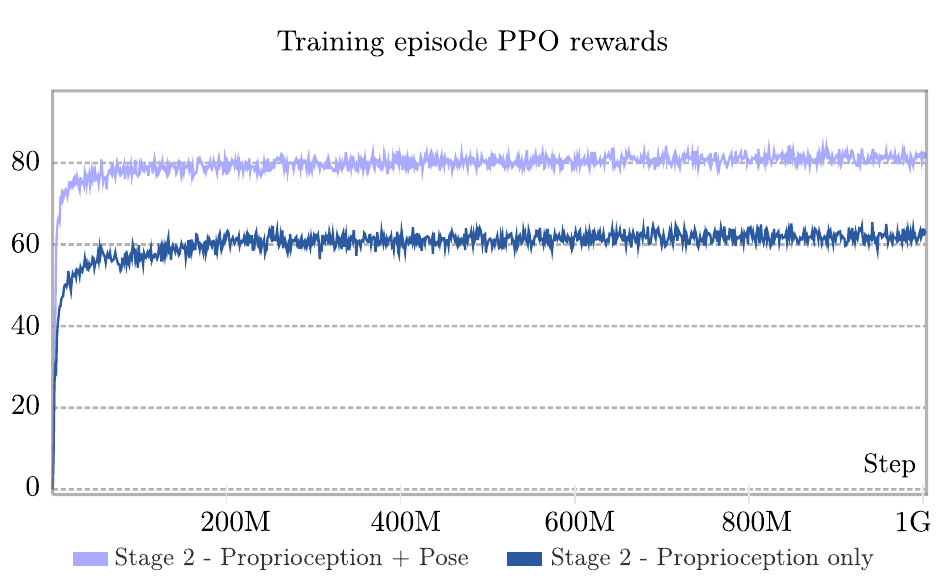}
        \label{fig:proprio-pose-vs-proprio-sim-reward}
    \end{subfigure}
    \hfill
    \begin{subfigure}[b]{0.48\textwidth}
        \centering
        \includegraphics[width=0.9\linewidth]{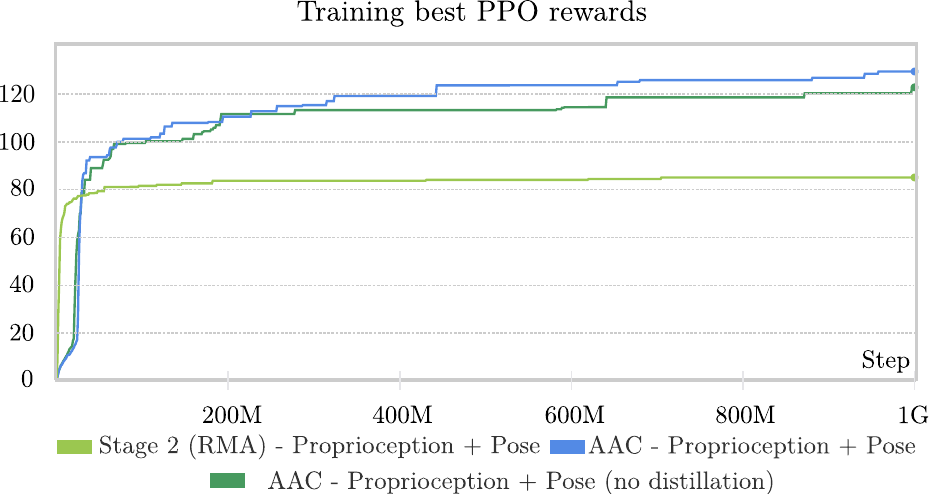}
        \label{fig:rma_aac_no_distillation}
    \end{subfigure}
    \caption{(a) Stage 2 distillation improves significantly when object pose is provided in addition to proprioception. (b) AAC (blue) trained in a single stage outperforms the RMA distillation which requires two stage training}
    \label{fig:sim-results}
    \vspace{0.5em}
    \begin{minipage}[t]{0.48\textwidth}
        \centering
        \setlength{\tabcolsep}{4pt}
        \resizebox{\linewidth}{!}{%
        \begin{tabular}{lrrrr}
        \toprule & & \multicolumn{3}{c}{$z$-axis} \\
        Method & Input modalities & RotR $\uparrow$ & TTF $\uparrow$ & RotP $\downarrow$ \\
        \midrule
        \textbf{Oracle} & & \textbf{159.4} & \textbf{0.89} & \textbf{31.86} \\
        \cmidrule(r){1-2} \cmidrule(lr){3-5}
        RMA~\cite{qi2022hand} & Proprioception & 139.0 & 0.79 & 28.53 \\
        RMA~\cite{qi2022hand} & + Pose & 153.1 & 0.86 & 31.09 \\
        \cmidrule(r){1-2} \cmidrule(lr){3-5}
        AAC & Proprioception & 141.9 & 0.80 & 27.83 \\
        AAC & + Pose & \textbf{168.6} & \textbf{0.88} & \textbf{26.5} \\
        \bottomrule
        \end{tabular}
        }
        \captionof{table}{\textbf{Simulation baseline comparison.} AAC with latent supervision significantly outperforms existing benchmarks.\vspace{-2em}}
        \label{tab:rma_aac_no_distillation_sim}
    \end{minipage}
    \hfill
    \begin{minipage}[t]{0.48\textwidth}
        \centering
        \setlength{\tabcolsep}{4pt}
        \resizebox{\linewidth}{!}{%
        \begin{tabular}{llc}
        \toprule 
        & & {Avg. Rotation Error} \\
        Pose parameterization & Distillation modality & Rad $(\downarrow)$ \\
        \midrule
        Absolute object pose & Proprioception & \pms{0.43}{0.11} \\
        Absolute object pose & Proprioception + Tactile & \textbf{\pms{0.21}{0.03}}\\ 
        \midrule
        Relative object pose & Proprioception & \pms{0.95}{0.09} \\
        Relative object pose & Proprioception + Tactile & \textbf{\pms{0.26}{0.05}} \\
        \bottomrule
        \end{tabular}
        }
        \captionof{table}{\textbf{Orientation reconstruction.} Tactile feedback enables superior recovery of both absolute and relative poses.\vspace{-2em}}
        \label{tab:pose-reconstruction}
    \end{minipage}
\end{figure*}

\subsection{Privileged sensors improve in-hand rotation performance in simulation}\label{subsec:sim_results}
We first verify that privileged sensors significantly boost simulation in-hand rotation performance. \cref{fig:sim-results}(a) shows that incorporating object pose alongside proprioception yields higher rewards during distillation than the proprioception-only baseline. Further, we evaluate performance under randomized physics properties (\cref{tab:rma_aac_no_distillation_sim}) using metrics from \cite{qi2022hand}: (1) \textbf{RotR}: \emph{Rotation reward} ($\omega \cdot k$) (radians), (2) \textbf{TTF}: \emph{Time to Fall} (second), and (3) \textbf{RotP}: \emph{Undesired rotation penalty} (cm). As expected, privileged information improves the policy across all evaluation criteria.

We also compare AAC against RMA distillation. While RMA distillation learns faster initially, AAC particularly with online latent distillation (\cref{subsec:method-online-distillation}), eventually yields superior performance and real-world behaviors \cref{fig:sim-results}(b) . Due to AAC's superior performance (\cref{tab:rma_aac_no_distillation_sim}), we simplify the simulation phase to a single AAC training step.
 
\begin{figure*}[t]
    \centering
    \includegraphics[width=\linewidth]{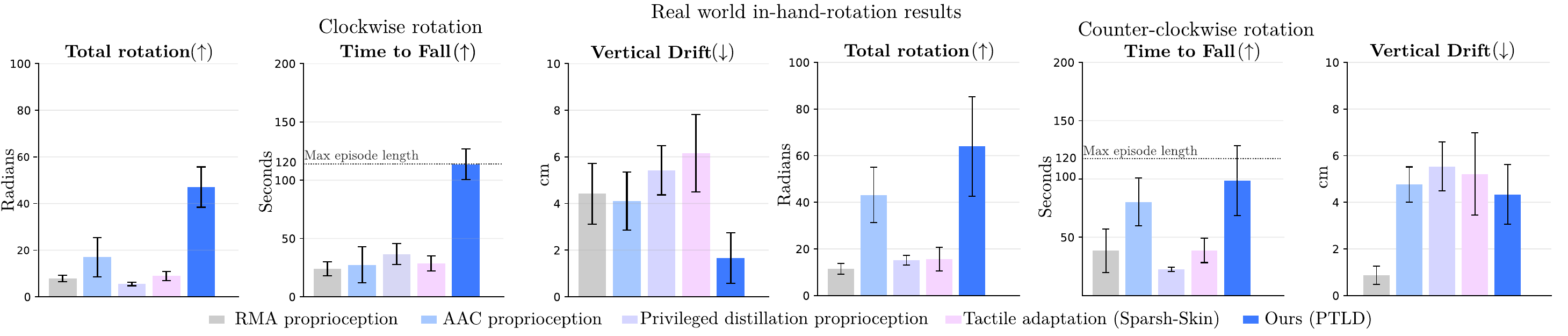}
    \begin{minipage}[b]{0.54\linewidth}
        \centering
        \includegraphics[width=\linewidth]{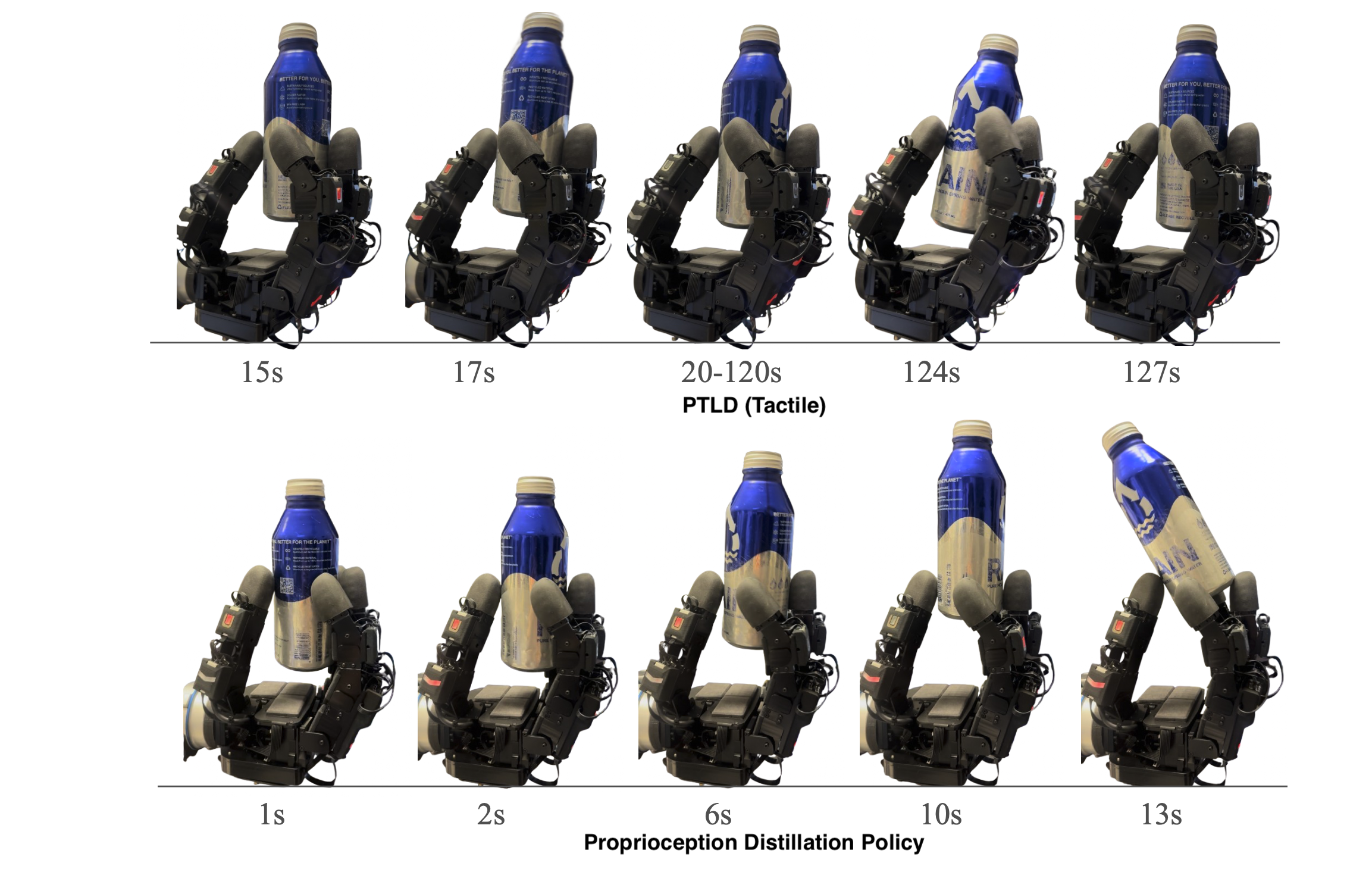}
  
    \end{minipage}
    \hfill
    \begin{minipage}[b]{0.44\linewidth}
        \centering
        \includegraphics[width=\linewidth]{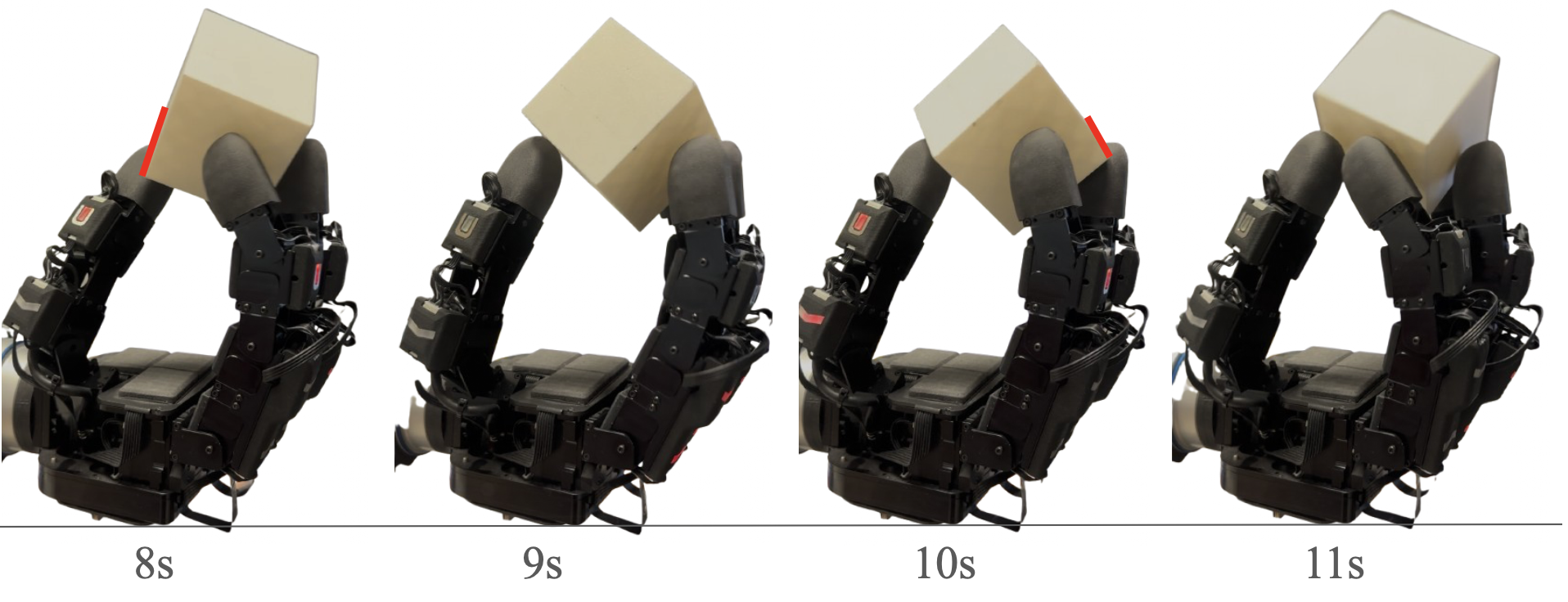}
        \vspace{0.5em}
        \resizebox{\linewidth}{!}{%
        \begin{tabular}{lrrrr}
        \toprule & & \multicolumn{3}{c}{$z$-axis} \\
        Method & Input modalities & RotR $\uparrow$ & TTF $\uparrow$ & RotP $\downarrow$ \\
        \midrule
        AAC & Proprio+Pose & \pms{17.3}{8.8} & \pms{0.33}{0.15} & \pms{5.5}{3.6} \\
        Proprio dist. & Proprio & \textbf{\pms{0.5}{0.5}} & \pms{0.11}{0.03} & \pms{1.3}{0.9} \\
        Tactile dist. & Tactile & \pms{4.5}{2.3} & \textbf{\pms{0.34}{0.21}} & \textbf{\pms{50.8}{28.1}} \\
        \bottomrule
        \end{tabular}
        }
        \captionof{table}{\ourmethod significantly improves rotation for cube object}
        \label{tab:cube-ptld}
    \end{minipage}

    \caption{\textbf{(top)} Real-world in-hand rotation (10 trials, 3 objects). \ourmethod significantly outperforms all baselines. \textbf{(bottom-left)} Qualitative recovery behavior: when the object is pushed above stable height at 17s, \ourmethod loosens its grip to allow the object to settle, then pushes it back up when it drops at 124s, sustaining rotation for over 220s. In contrast, the proprioception-only baseline fails to recover and drops the object at 13s. \textbf{(bottom-right)} \ourmethod on a cube-shaped object. Red highlights indicate edge contact; once detected, the policy adjusts finger gait to transition from unstable edge contact to stable face contact. For instance, at 8s the thumb contacts the left edge; by 9s the policy actively repositions the thumb onto the face, restoring a stable grasp.\vspace{-2em}
    }
    \label{fig:real-world-baselines-vs-ours}
\end{figure*}

\subsection{\ourmethod improves policy in-hand rotation robustness in the real world}\label{subsec:real_results}
We evaluate \ourmethod on hardware against the following baselines: (1) \textbf{RMA Proprioception}\cite{qi2022hand}, two-stage distillation deployed with proprioception only (2) \textbf{AAC Proprioception}, single-stage AAC policy deployed with proprioception only. (3) \textbf{Proprioceptive Distillation}, \ourmethod's pipeline with proprioception substituted for tactile at the real-world distillation step, isolating the contribution of tactile sensing; and (4) \textbf{Tactile Adaptation}\cite{higuera2025tactile} using Sparsh-skin\cite{sharma2025selfsupervised} representations, which distills into tactile from a proprio-only teacher, isolating the contribution of the privileged teacher. Each baseline is evaluated across 10 trials on two cylindrical objects on Total Rotation (radians), Time to Fall (s), and Vertical Drift (cm).

As shown in \cref{fig:real-world-baselines-vs-ours}, \ourmethod significantly outperforms all baselines in rotation and TTF. The tactile student in \ourmethod captures \textit{dynamic contact sequences} that proprioception alone cannot. Specifically, during slip or upward push events, object pose changes while fingers continue their normal rotation gait, leaving proprioceptive signals ambiguous. Yet, the resulting subtle force distribution changes are directly captured by tactile signals. Table~\ref{tab:pose-reconstruction}  supports tactile’s sensitivity to pose changes, showing that tactile latents significantly outperform proprioception latents in encoding relative pose information. As a result, \ourmethod demonstrates adaptive finger-gaiting and sophisticated slip recovery behaviors inherited from the teacher (\cref{fig:real-world-baselines-vs-ours} (bottom-left)). While RMA shows lower vertical drift in counter-clockwise trials, it suffers from significant object tilting and off-axis rotation, which \ourmethod avoids.

To demonstrate object diversity, we additionally evaluate \ourmethod on cube rotation. For irregular geometries where Aruco markers are impractical due to occlusion, we utilize FoundationPose~\cite{foundationposewen2024} for markerless object pose estimation. As shown in~\cref{tab:cube-ptld}, \ourmethod significantly outperforms proprioceptive baselines, which fail almost immediately. We hypothesize that tactile signals enable the policy to distinguish between edge and face contact, enabling the policy to detect incipient instability and actively correct from unstable edge contact to stable face contact (\cref{fig:real-world-baselines-vs-ours} (bottom-right)). Joint angles are not rich enough to capture such contact geometry and learn corrections. Furthermore, joint angles are determined by both object pose and contact forces simultaneously, conflating the two and making pose estimation challenging. Tactile feedback provides a clean geometric signal for precise orientation estimation. 
Notably, \ourmethod outperforms the AAC privileged-sensor teacher policy. We attribute this to drift in FoundationPose pose estimates over time, which degrades the teacher policy but does not affect the distilled tactile policy. This highlights a key advantage of distillation: the student policy can surpass its teacher by operating on a more robust sensing modality at deployment.\vspace{-1em}

\subsection{Tactile information enhances object state estimation}\label{subsec:tactile-reconstruction-result}
\begin{figure*}
  \centering
  \begin{minipage}[b]{0.58\textwidth}
    \centering
    \includegraphics[width=\linewidth]{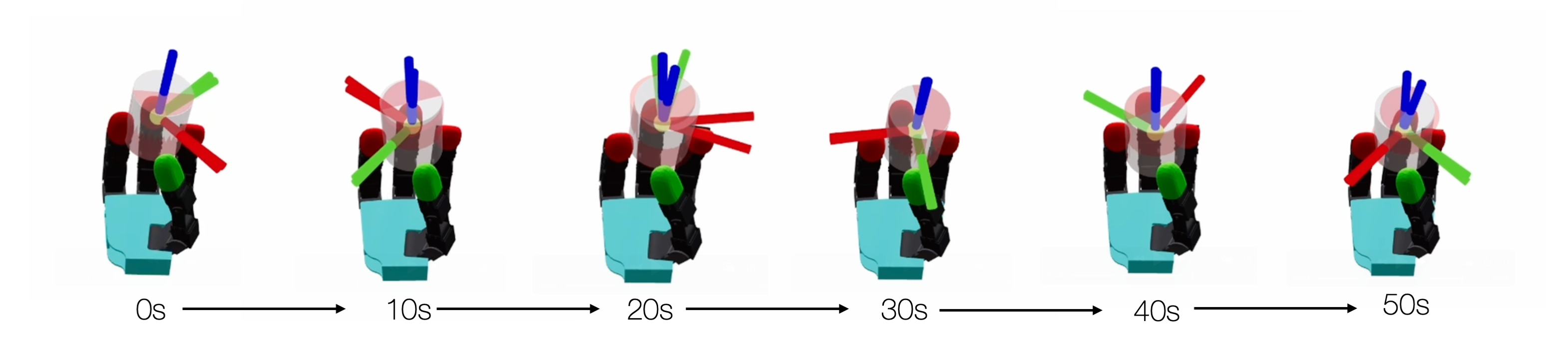}
    \caption{Object pose reconstruction from tactile latents. Red and gray cylinders denote predicted and ground-truth poses, respectively, composed per second.}
    \label{fig:tactile_orientation_pred}
  \end{minipage}
  \hfill
  \begin{minipage}[b]{0.38\textwidth}
    \centering
    \resizebox{\linewidth}{!}{%
    \begin{tabular}{lcc}
    \toprule & \multicolumn{2}{c}{In-hand reorientation} \\
    Architecture & $N_\text{goals}~(\uparrow)$ & TTF (s) $(\uparrow)$ \\
    \midrule
    Transformer \textbf{(proprio)} & \pms{2.1}{2.14} & \pms{10.99}{10.15} \\
    Transformer \textbf{(tactile+proprio)} & \textbf{\pms{3.3}{1.55}} & \textbf{\pms{13.42}{10.18}} \\
    \bottomrule
    \end{tabular}
    }
    \vspace{1em}
    \captionof{table}{In-hand reorientation (10 trials). Success decreases significantly without tactile feedback.}
    \label{tab:in-hand-reorientation}
  \end{minipage}
\end{figure*}

\vspace{-0.5em}
To analyze information overlap between privileged and tactile policies, we measure the object orientation encoded in tactile latents. We trained a 6D rotation decoder to predict absolute ($\mathbf{R}_t^W$) and relative ($\mathbf{R}_t^{t-H}$) object poses from student encoder rollouts. As shown in Table~\ref{tab:pose-reconstruction}, tactile information significantly reduces prediction error for both parameterizations. Proprioception-only encoders struggle particularly with relative pose; we attribute this to the ambiguity of finger-gait patterns when unobserved slippage occurs, which tactile feedback disambiguates via direct contact signals. Absolute pose estimation results are visualized in Figure~\ref{fig:tactile_orientation_pred}.\vspace{-0.5em}

\subsection{Tactile object In-hand reorientation}\label{subsec:reorientation-results}

The in-hand reorientation policy (\cref{subsec:reorientation}) benchmarks \ourmethod's capacity for high-precision state estimation. We observe here that temporal-convolutional encoders often collapse into uni-directional gaiting, however, the Autoregressive Transformer captures the long-range dependencies required for multi-directional reorientation. Performance is quantified by goals reached ($N_{\text{goals}}$) and Time to Fall (TTF). As shown in Table~\ref{tab:in-hand-reorientation}, success rates drop significantly without tactile input, which proved essential for real-time slip detection and compensatory adjustments. 

\section{Conclusion}\label{sec:conclusion}\vspace{-0.5em}

In this paper, we presented \ourmethod, a framework for learning tactile dexterous manipulation policies without the need for accurate tactile simulation. Our core insight is to move beyond the traditional zero-shot sim-to-real constraint. Instead, we execute policies in real-world instrumented setups using privileged sensors, then distill these rollouts into tactile policies by aligning the implicit state latents of the privileged sensor policy with those of the tactile policy. We demonstrated the efficacy of \ourmethod on complex tasks, including in-hand rotation and reorientation, achieving significant performance gains in both. While this work focuses on tactile sensing, our methodology is general-purpose and provides a blueprint for learning perceptive policies across other modalities, such as vision.

\textbf{Limitations:} While \ourmethod offers a novel streamlined approach to training policies with sensing modalities that are difficult to simulate, we identify the following limitations:
\begin{itemize}[itemsep=-2pt,topsep=-3pt,leftmargin=5mm]
\item \textbf{Information Overlap \& Asymmetry:} \ourmethod relies on distillation from a privileged sensor policy. Success depends on the information overlap between the privileged and deployment sensors. For instance, while a tactile sensor captures rich dynamics directly via contact patches and forces, object poses capture the \emph{effect} of forces indirectly as kinematic data. Selecting a privileged sensor that minimizes information asymmetry is critical for high-fidelity distillation. 

\item \textbf{Privileged Sensor Noise Floors:} All real-world sensors possess inherent noise. Because our distillation source is a policy trained in simulation, it must be robust to the specific noise profiles of real-world privileged sensors. High noise in object pose estimation, for instance, sets a performance ceiling on the deployed policy. High-precision systems like motion capture can help mitigate this. 

\item \textbf{Instrumented Setup Requirements:} The requirement for an instrumented setup (e.g., external cameras or trackers) to provide privileged state information during training (distillation) may limit rapid application of this method to completely unstructured or "in-the-wild" environments.
\end{itemize}

\clearpage
\acknowledgments{We thank Haozhi Qi, Carolina Higuera, Guanya Shi and Chaoyi Pan for initial discussions and feedback about the project. We are also grateful to Zilin Si, Changhao Wang, Youngsun Wi, Unnat Jain and Jay Karhade for feedback on the paper draft.  }


\bibliography{references}  
\clearpage




\begin{appendix}

\section{Hardware and Implementation details}

\subsection{Privileged sensor cell}
For all experiments, we use the Allegro hand sensorized with Xela uSkin~\cite{tomo2018xela}, attached to a Franka Panda robot arm. 18 Xela uSkin sensing pads cover the Allegro hand 
amounting to a total of 368 individual sensors. We use the 3-axis raw sensor measurements over processed force measurements as force measurements contain significant hysteresis and lag. A baseline signal with no contact is additionally collected (over 2 minutes) and subtracted from the raw tactile measurements before data collection each time.

To deploy the privileged sensor policies for latent dataset collection, we instrument the real world robot cell with 4 Realsense D435i/D435 RGBD cameras which view the in-hand manipulation area. These cameras are calibrated jointly and track an Aruco marker attached to the object being manipulated to produce reliable multi-view pose estimate that is refined via Pose Graph Optimization (PGO)~\cite{gtsam}. We assume known shapes for the objects that are being deployed.

\subsection{Implementation details} 
We use IsaacGym~\cite{makoviychuk2021isaac} as our simulator. Once an offline dataset is collected in the real-world cell, the tactile encoder is trained via supervised learning. We use AdamW optimizer, with a LR of $1\mathrm{e}{-4}$. We use ROS2 for communication between the different robot processes, and achieve a real-time policy deployment rate of $\sim20$Hz for both in-hand rotation and in-hand reorientation.
\vspace{-1em}

\section{Grasp generation}
For both the policies we consider in this paper, we generate a grasp cache tuned for the Xela hand, which provides initial stable grasps as starting points for the episode. Inspired from~\cite{qi2022hand}, to generate the stable grasp poses, first, we initialize the robot hand (Xela hand) to a canonical pose and randomize the joint configuration within set limits, while the object is initialized slightly above the robot hand. Once the simulation proceeds, we retain the grasps if the object is held stably after 50 simulation steps.

\section{In-hand rotation}

\subsection{Object diversity}
\begin{figure}
    \centering
    \includegraphics[width=0.5\linewidth]{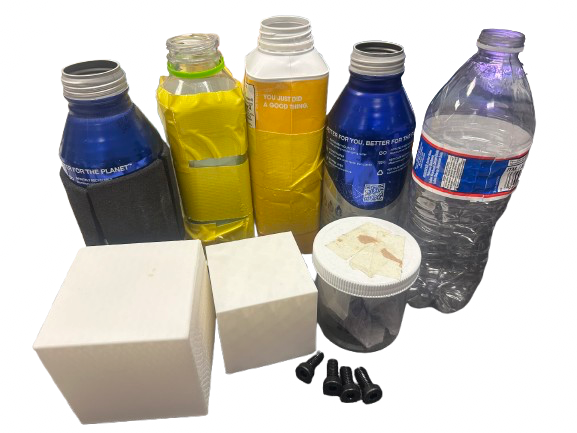}
    \caption{objects used in data collection and policy evaluation}
    \label{fig:objects}
\end{figure}

Figure~\ref{fig:objects} shows the set of objects used for data collection and policy evaluation. We use four cylindrical objects with radii of 30 mm, 31.75 mm (×2), and 33.4 mm; heights of 70 mm, 178 mm (×2), and 200 mm; and varying surface frictions. In addition, we use two square bottles with side lengths of 54 mm and 60 mm, and heights of 187 mm and 194 mm. For object generalization experiments, we used 2 3D-printed cubes with edges of 64mm and 80mm.

The object masses range from 22 g to 90 g (22 g, 24 g, 30 g, 46 g, 87 g, and 90 g). To further vary the mass during data collection and evaluation, we insert additional weights of 18 g and 36 g into the bottles. Each weight piece (black nut) has a mass of 9 g.

\subsection{Data collection}
For each tactile policy, we collect approximately 180 minutes of real-world rotation trajectory data. Specifically, in each DAgger iteration, we gather 30 trajectories of 2 minutes each, and perform 3 rounds of DAgger in total to obtain the final policy.
Within each batch of 30 trajectories, we systematically vary object properties—including shape, mass, texture, and weight—to ensure balanced coverage and reduce bias in the dataset. This data collection strategy allows us to capture a wide range of feasible object states and contact conditions prior to object dropping, leading to a more robust tactile policy. 

\section{In-hand reorientation}
\subsection{Reward details}
\begin{table*}[t]
\centering
\small
\begin{minipage}[t]{0.48\textwidth}
\centering
\setlength{\tabcolsep}{4pt}
\renewcommand{\arraystretch}{1.1}
\resizebox{\linewidth}{!}{%
\begin{tabular}{lc}
\toprule  
\textbf{Reward} & \textbf{Scale}\\
\midrule
$r_{\text{goal}} \triangleq \frac{1}{d(\Rv_t^\text{object}, \Rv_t^\text{goal}) + \epsilon}$ & $2.0$ \\
$r_\text{success} \triangleq (1~\text{if}~d(\Rv_t^\text{object}, \Rv_t^\text{goal}) \leq \delta~\text{else}~0)$  & $5.0$    \\
$r_\text{streak} \triangleq \frac{N_\text{success}}{N_\text{max\_success}}$ & $2.0$  \\
$r_\text{contact} \triangleq \sum_i (C_i > \delta_\text{contact})$  & $0.1$ \\
$r_\text{position} \triangleq \| p_t - p_0 \|$ & $0.05$ \\
$r_\text{finger\_pose} \triangleq \| q_t - q_0 \|$ & $-1.0$ \\
$r_\text{fingertip\_object} \triangleq \sum_i \| p_{\text{fingertip}_i} - p_t \|$ & $-0.2$ \\
\bottomrule
\end{tabular}
}
\captionof{table}{Reward function for target in-hand reorientation.}
\label{table:task_rewards}
\end{minipage}
\hfill
\begin{minipage}[t]{0.48\textwidth}
\centering
\setlength{\tabcolsep}{4pt}
\renewcommand{\arraystretch}{1.1}
\resizebox{\linewidth}{!}{%
\begin{tabular}{llc}
\toprule  
\textbf{Reward} & \textbf{Scale}\\
\midrule
$r_\text{angular\_velocity}$ & $\triangleq \| \frac{(\Rv_t \cdot \Rv_{t-1}^\top)}{\Delta t} \|$ & $-0.05$ \\
$r_\text{acceleration}$ & $\triangleq \| \dot{q_t} - \dot{q_{t-1}} \|$ & $-0.005$ \\
$r_\text{action}$ & $\triangleq \| a_t \|$ & $-0.005$ \\
$r_\text{action\_rate}$ & $\triangleq \| a_t - a_{t-1} \|$ & $-0.2$ \\ 
$r_\text{joint\_limit}$ & $\triangleq \max(q_\text{lower} - q_t, 0) + $ & \\
& \phantom{$\triangleq$} $\max(q_t - q_\text{upper}, 0) $ & $-0.1$ \\
$r_\text{object\_velocity}$ & $\triangleq \| \frac{p_t - p_{t-1}}{\Delta t} \|$ & $-1.0$ \\
$r_\text{torque}$ & $\triangleq \| \tau \|$ & $-0.5$ \\
$r_\text{work}$ & $\triangleq \| \tau \cdot \Delta q \|$ & $-4.0$ \\
$r_\text{timeout}$ & $\triangleq t \geq T$ & $-1.0$ \\
$r_\text{alive}$ & $\triangleq t - t_0$ & $-0.01$\\
\bottomrule
\end{tabular}
}
\captionof{table}{Motion and energy penalties for reorientation.}
\label{table:penalties}
\end{minipage}
\end{table*}

\paragraph{Reward} We train the policy with a reward mixture (Table~\cref{table:task_rewards}) that balances goal reaching with natural finger gaits. Primary components include rotational distance ($r_\text{goal}$), success bonuses ($r_\text{success}$), and a streak bonus ($r_\text{streak}$) to encourage sequential task completion. To ensure stability, $r_\text{contact}$ and $r_\text{position}$ incentivize fingertip contact and object centering. We apply motion and energy penalties for smoothness; crucially, penalizing deviations from the initial finger pose ($r_\text{finger\_pose}$) prevents fingers from curling into unrecoverable configurations and promotes gait-like behavior. In addition to the rewards described in ~\cref{table:task_rewards}, we also use additional reward shaping terms to ensure that the policy results in natural gaits. We detail the rewards in~\cref{table:penalties}
\paragraph{Reset}
Since we expect the policy to reach several goals in an episode, we design the reset strategy accordingly. 
We start the episode with a stable grasp sampled from a grasp set as commonly implemented, and reset the goal orientations multiple times within an episode when the object orientation reaches the goal orientation ($\leq \delta$). Additionally, we use a $z$-height threshold to reset the episode when the robot hand drops the object from its fingers or the object slips to unrecoverable states. Finally, we also use the standard episode reset after a fixed number of simulation steps. 

\subsection{Additional results of in-hand reorientation policy}

\begin{figure*}
    \centering
    \includegraphics[width=\linewidth]{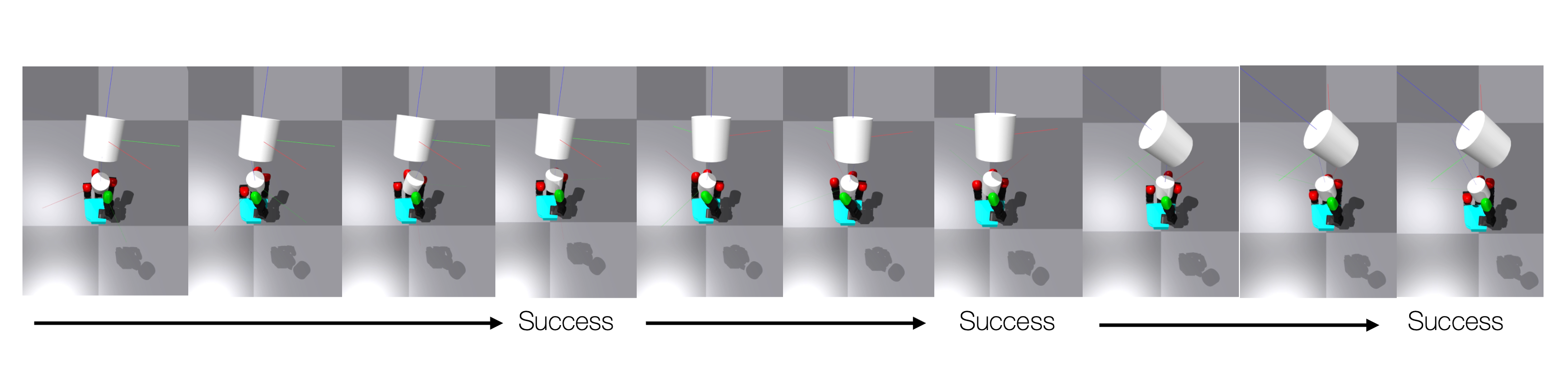}  
    \caption{Visualization of simulation in-hand reorientation}\label{fig:in-hand-reorientation-sim-appendix}
\end{figure*}

\begin{figure*}
    \centering
    \includegraphics[width=\linewidth]{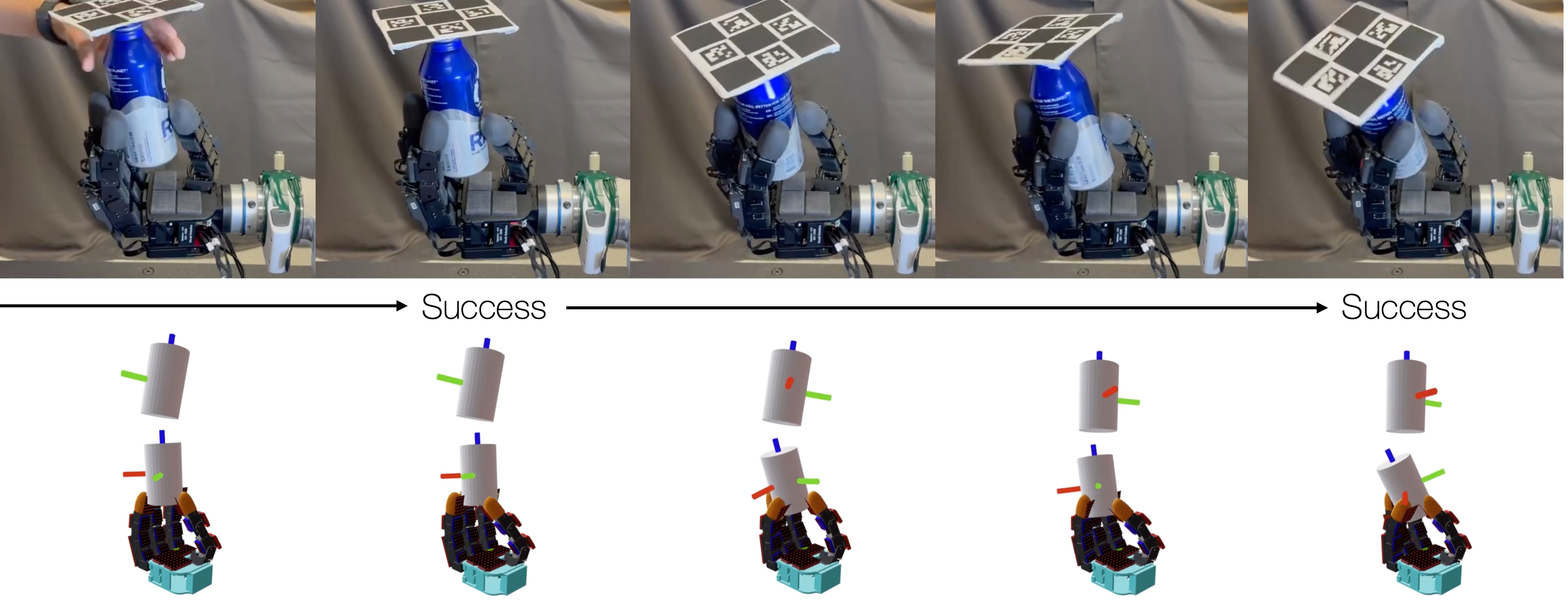}  
    \caption{Visualization of real world tactile in-hand reorientation. (Marker only used for goal-reaching evaluation. The policy only takes tactile and/or proprioception signals, it does not take object pose as input.) }\label{fig:in-hand-reorientation-real-appendix}
\end{figure*}

In~\cref{fig:in-hand-reorientation-sim-appendix} we visualize a roll out of the AAC trained policy in simulation showing continuous goal reaching behaviors. In simulation the policy is provided access to current object orientation as well as goal orientation in addition to proprioception. Furthermore in simulation the policies are trained for a wider distribution of goal poses ($\sim 40^\circ$ about the $z$-axis), and the success threshold is tighter ($= 20^\circ$). Similarly in~\cref{fig:in-hand-reorientation-real-appendix} we visualize a real world roll out of the tactile in-hand-reorientation. We find that the tactile policy is able to manipulate objects reaching \textbf{multiple goal }poses within a single deployment. A demo is also additionally provided in the video submission. It must be noted that the marker is only used to evaluate whether a goal pose has been reached, and the policy does not take current object pose as input. In the real world we relax the success threshold to $\sim 30^\circ$. 

\end{appendix}

\end{document}